\documentclass{article}
\usepackage[utf8]{inputenc}
\usepackage{comment}
\usepackage[dvipsnames]{xcolor}
\usepackage{soul}
\usepackage{multicol}
\usepackage{tabularx}
\usepackage{subcaption}
\usepackage{placeins}
\usepackage{pbox}
\usepackage{amsfonts}
\usepackage{amsmath}
\usepackage{graphicx}
\usepackage{enumitem}
\usepackage{pifont}
\usepackage{makecell}
\usepackage{multirow}
\usepackage{hyperref}
\usepackage[switch]{lineno}
\usepackage[margin=1in]{geometry}
\usepackage{authblk}

\newcommand{\cmark}{\ding{51}}%
\newcommand{\xmark}{\ding{55}}

\newcommand{\TP}{\mathit{TP}}
\newcommand{\TN}{\mathit{TN}}
\newcommand{\FP}{\mathit{FP}}
\newcommand{\FN}{\mathit{FN}}
\newcommand{\PPR}{\mathit{PPR}}
\newcommand{\NPR}{\mathit{NPR}}
\newcommand{\FPR}{\mathit{FPR}}
\newcommand{\EO}{\mathit{EO}}
\newcommand{\Sim}{\mathit{Sim}}

\begin{document}

\title{
Streamlining models with explanations in the learning loop\vspace{1cm}
}

\author[1]{Francesco Lomuscio}
\author[2]{Paolo Bajardi}
\author[2]{Alan Perotti}
\author[1]{Elvio G. Amparore}
\affil[1]{Università di Torino, Turin, Italy}
\affil[2]{CENTAI Institute, Turin, Italy}
\setcounter{Maxaffil}{0}


\date{\vspace*{-5mm}mailto: amparore@di.unito.it, alan.perotti@centai.eu}

\maketitle

\thispagestyle{empty} 
\pagestyle{plain} 

\begin{abstract}
Several explainable AI methods allow a Machine Learning user to get insights on the classification process of a black-box model in the form of local linear explanations. 
With such information, the user can judge which features are locally relevant for the classification outcome, and get an understanding of how the model reasons. 
Standard supervised learning processes are purely driven by the original features and target labels, without any feedback loop informed by the local relevance of the features identified by the post-hoc explanations.
\\
In this paper, we exploit this newly obtained information to design a feature engineering phase, where we combine explanations with feature values. 
To do so, we develop two different strategies, named \textit{Iterative Dataset Weighting} and \textit{Targeted Replacement Values}, which generate \textit{streamlined models} that 
better mimic the explanation process presented to the user.
We show how these streamlined models compare to the original black-box classifiers, in terms of accuracy and compactness of the newly produced explanations.
\end{abstract}

\section{Introduction}
\label{sec:intro}

Traditionally, supervised Machine Learning (ML) is performed by training a model instance from a labeled input dataset. 
In particular, modern ML models are typically \textit{black boxes} - that is, models that do not intrinsically share a human-comprehensible description of their inner decision logic.

Explainable Artificial Intelligence (XAI) allows a user (be it an engineer or a domain expert) to perform various kinds of a-posteriori attributions to black-box models, to partially explicate how these model take their decisions in a human-comprehensible format~\cite{bib:guidotti}.
Depending on the data type, ML task and analysis technique, explanations can be computed and provided in different ways, such as lists of rules, decision trees, weights, prototypes, etc. 
Of particular interest is the \textit{feature attribution}~\cite{lundberg2017unified} approach, which represents the explanation to any given sample $x$ as a \textit{local linear explanation} (LLE).
The LLE of $x$ assigns a weight to each feature of $x$, representing its relative importance in the classification outcome.

The typical exploitation of a LLE is to focus the attention on the (locally) most relevant features, which implicitly presents the idea that the model based its decision essentially on such features.
This simplification is driven by the need to present to the user/developer an amount of information that is cognitively manageable; however, this process might in fact introduce a mismatch between the machine decision and its corresponding human understanding.
For instance, a XAI technique might highlight a subset of locally relevant features for a given data point, but these features alone might even produce a different output when fed to the same trained ML model.
It is therefore interesting to assess whether the most relevant features presented to the user are actually enough to preserve the classification task, i.e. reconciling both the machine and the human learning processes.

Therefore, our research question (RQ) is: 
\begin{equation}\label{eq:RQ}
\begin{array}{c}
\text{\it How can local linear explanations be exploited in a ML workflow so that the trained model} \\
\text{\it is more coherent to the human understanding of the explanation process?} \\
\end{array}
\end{equation}

In this paper we present an experimental setup of two novel feature-engineering techniques, where a \textit{streamlined black-box model} $h$ is trained over a dataset where only the locally-relevant features are kept, while the others are masked away.
By doing so, the resulting streamlined model $h$ surpasses the black-box model in mimicking the psychological process of the user receiving an explanation, focusing on the most relevant features, 
and taking decisions upon these reduced inputs.
Our target is therefore to test whether the new model really learned to classify upon the suggested subset of features without compromising classification accuracy and model fairness. 
The streamlined model is still a black box, but the extracted explanations are more compact, and therefore more comprehensible.
The two presented methodologies are called \textit{Iterative Dataset Weighting} (IDW) and \textit{Targeted Replacement Values} (TRV).
IDW is designed as an iterative procedure that directly combines the dataset values with their importance, in order to lower towards zero the features that are not relevant according to the LLEs.
A specific pre-processing is also introduced to reduce the possibility that the zero value corresponds to a highly relevant value.
TRV identifies a neutral value for each feature, and performs value replacements to keep only a predefined cognitive budget of $K$ relevant feature per sample.

For each methodology, a workflow that generates the streamlined models is proposed and assessed experimentally on a set of 4 benchmark tabular datasets.
The experiments show that the proposed methodologies produce a trade-off between the baseline accuracy and the complexity of the streamlined model logic, computed in terms of \textit{explanation compactness} and \textit{similarity} to a common global explanation.

The rest of the paper is organised as follows: \ref{sec:relatedworks} presents background and related work, \ref{sec:contributo} introduces the two proposed methodologies and the dataset processing,
\ref{sec:res} describes the experimental setup and results, and \ref{sec:conclusions} concludes the paper with final considerations and directions for future work.

\FloatBarrier

\section{Related Works}
\label{sec:relatedworks}

In recent years, several works have been dedicated to the concept of AI explanation. 
There are models that are inherently understandable by humans. For example, decision trees provide an easily interpretable explanation as the path from the root to the leaf that classifies a sample~\cite{decisiontree}, as well as rule-based models~\cite{associationrules} or linear regressors.
Most of the ML models in use today, however, are not \textit{understandable by design}, and exploit complex mathematical architectures that are hard to interpret directly.
Models of that kind include neural networks, random forests, and others, and are usually called \textit{black-box} models for this reason.
Interpreting black-box models can still be done by using techniques to extract various representations (such as prototypes, heatmaps, etc.) that somehow capture relevant parts of the model classification process~\cite{DBLP:journals/corr/abs-1909-12072}.
These representations can pertain to single data points (\textit{local explanations}) or concern the behavior of the black-box model as a whole (\textit{global explanations}).

A common representation\,\cite{covert2021explaining} for local explanations of black-box models is the \emph{Local Linear Explanation} (LLE), which provides a coefficient for every input feature that represent the importance of such feature in the classification outcome.
LLEs are particularly suited for tabular data, but can be generalized to cover a broader class of data formats. 
Different methodologies exists to compute LLEs, such as LIME~\cite{ribeiro2016why}, DeepLift~\cite{shrikumar2017just} or SHAP~\cite{lundberg2017unified}. 
In this paper we will focus on SHAP, which is briefly summarized in the section \ref{sec:shap}, as the methodology to compute LLEs.

\subsection{SHapley Additive exPlanations (SHAP)}
\label{sec:shap}
SHAP~\cite{lundberg2017unified} is the state-of-the-art methodology to compute LLEs that we will use throughout the paper.
In particular, we focus on the \textit{SHAP KernelExplainer method} (hereafter \textit{SHAP explanation method}, for brevity), which is the core implementation of the Shapley algorithm that does not make any assumptions on the nature of the black-box model $f_0$.

The \textit{SHAP explanation method} takes in input a model $f_0$ and a sample $x$, and computes as a result a vector of SHAP scores $\phi$, which can be interpreted as the relative importance of that feature in the classification process of a model $f_0(x)$.
In principle this explanation method is defined on binary features only, but it can be extended to real values ones.
In order to compute the scores, SHAP needs to evaluate the black-box $f_0$ on a subset of features $S \subseteq F$ of $x$. To do so, SHAP adopts the concept of a \textit{background set} $B$, which is a set of samples whose entries are used to replace the missing features of $S$.
Let $f_{S,b}(x)$ denote the evaluation of $x$ where only the features of $S$ from $x$ are kept, while the other features $F \setminus S$ are taken from the values of the background sample $b \in B$,
and let $f_{S}(x) = \frac{1}{|B|} \sum_{b \in B} f_{S,b}(x)$.

An explanation of a sample $x$ computed by SHAP is a vector $\phi=(\phi_0, \phi_1, \ldots, \phi_m)$, such that
\begin{equation}\label{eq:shapfx}
    f_0(x) = \phi_0 + \sum_{j=1}^m \phi_j    
\end{equation}
The score $\phi_j$ for feature $1 \leq j \leq m$ represents how much that feature $j$ contributes (positively or negatively) to the final classification value $f_0(x)$ with regard to a \textit{background score} $\phi_0$, defined as $\phi_0 = \frac{1}{|B|} \sum_{b \in B} f_0(b)$.
Each score $\phi_j$ can be either positive or negative, since it contributes to explain the value $f_0(x) - \phi_0$ which again can be either positive or negative.
The absolute value of a SHAP score $|\phi_j|$ can be interpreted as a \textit{local feature importance} for feature $j$ of sample $x$.
Given a dataset $D$, let $E$ be the matrix of all SHAP scores of all samples of $D$, such that $E_{ij}$ is the SHAP score of feature $j$ of sample $i$. Note that $\phi_0$ does not need to be represented in $E$, being an immutable value.

\begin{table}[t]
\centering
\begin{tabular}{ | c | l | }
 \hline \textbf{Symbol} & \textbf{Meaning} \\ \hline 
  $f_0$ & baseline black-box model \\ \hline
  $F$ & the set of $m$ features of the dataset \\ \hline
  $D$ & dataset matrix with $n$ samples (size $n \times m$)\\ \hline
  $x_j$ & the value of the data point $x$ for the feature $j$ \\ \hline
  $B$ & matrix of $k$ background data points (size $k \times m$) \\ \hline
  $\phi$ & explanation vector for a single data point $x$ \\ \hline
  $E$ & matrix of all explanation vectors $\phi$ (size $n \times m$) \\ \hline
  $K$ & the cognitive load allowed for a linear explanation \\ \hline
  $h$ & streamlined black-box model \\\hline
  $R$ & replacement values (size $1 \times m$) \\\hline

\end{tabular}
\caption{Summary of symbols used in the paper.}
\label{tab:notation}
\end{table}


\subsection{Learning from explanations}
\label{sec:posthoc}

Explanations of black-box classifiers can be seen as artifacts generated to help a human subject in understanding the internal process of a trained model. 
Since the model has to be trained beforehand, these explanations are \textit{post-hoc}.
It is therefore relevant that explanations are kept psychologically compact and convincing, and this may require some cognitive load constraint to be defined and enforced~\cite{abdul2020cogam}.

Inspired by well-established contributions in cognitive psychology research and education~\cite{chi1989self,chizhik2001equity,bobek2016creating},
a different but compelling idea consists in reusing post-hoc explanations generated by a model to improve the learning process.
In~\cite{hase2021models} the conditions that make post-hoc explanations useful to improve a model are studied. The proposed approach is based on the use of past explanations for predicting unknown data points.
In~\cite{bib:WT5} the classification task is extended to generate the explanations together with the target, thus making the black-box model (to some extent) self-explanatory.
The work in~\cite{bib:reflectiveNet} explores the idea of combining the learning process of a deep neural network model with the explanations (generated by the Grad-CAM method\,\cite{selvaraju2017grad}) for multiple outcome classes, in order to let the model learn from the  explanation for every possible outcome.
Another possible use of explanations consists in learning a secondary model that can be used to fine tune a decision process, like in~\cite{cleger2014learning} for changing the output of a recommendation system.

Recently, a self-explanation module has been proposed to complement existing deep learning pipelines to incorporate \textit{ante-hoc} explanations~\cite{sarkar2022framework}. The suggested methodology is able to provide explanations for model decisions in terms of concepts for an individual input image and can handle different levels of supervision.

Classification quality improved through the use of explanation is studied in~\cite{ye2021can}: in that case, a sample and its explanation is passed to a \textit{calibrator model} to guess if the classification of a black-box model is correct.

In this paper we want instead to combine explanations generated from an initial model $f_0$ to perform a feature engineering of a dataset $D$, which is then used to train a streamlined model $h$ that should incorporate part of the knowledge injected in the engineering phase.

Post-hoc explanations techniques can also be valuable to implement more effective auditing processes~\cite{panigutti2021fairlens}. Understanding the logic behind wrong model decisions could be of paramount importance, especially when high stakes are in place.
Explaining a black-box can help finding mislabeled entries in massive datasets, like with the A-ClArC and P-ClArC methodologies~\cite{anders2020finding}. 

Explanations should also be subject to a validation process, to verify their reliability and algorithmic stability. A review of scores and techniques for that purpose can be found in~\cite{bib:LEAF}.

The notation used throughout the paper is summarized in Table~\ref{tab:notation}.

\FloatBarrier

\color{black}
\section{Methodologies}
\label{sec:contributo}



In this chapter, the two novel methodologies are introduced, namely \textit{Iterative Dataset Weighting} and \textit{Targeted Replacement Values}.
The purpose of these methodologies is to inject knowledge derived from the explanations about the logic of a black-box model $f_0$ into the training process of one (or more) new models, called \textit{streamlined models}.
To support these methodologies, a special dataset preparation is also introduced, with the purpose of using zero values as a way to encode non-relevant or average values.

\subsection{Data preparation and background values}

Datasets were preprocessed to prepare the raw data for ingestion into ML models:
all samples with missing values were removed, the datasets were then rebalanced using the SMOTE methodology~\cite{2002},
and all features were then normalized.
Numerical features were standardized using z-score normalization; therefore the value $0$ for a numerical feature corresponds to the expected value for that feature.
Binary categorical features were encoded using a one-hot scheme with $\{-1,1\}$ values.
All $N$-ary categorical features with $N$ possible values were transformed to $N$ binary categorical features.
By doing so, we can use the $0$ value to represent the lack of information about categorical variables.
The rationale behind this encoding is that we expect SHAP to assign lower explanation scores for features with values close to $0$, as they do not diverge from (uninformative) background values.



\label{sec:BG}
The computation of the Shapley values requires to remove players from coalitions, and SHAP mimics so by introducing the concept of feature masking: replacing a feature with a default value called background.
In principle the median value of each feature could be used, as in~\cite{shrikumar2017just}. 
Using $k$ background values improves the generality of the SHAP method\,\cite{backgroundShapFluctuations}, at the expense of increasing the computational cost by a factor of $k$.
As a balance between generality and cost, we have used K-means to obtain a set of $k=4$ background samples for each of the tested datasets.
Some of the tested datasets include subgroups of individuals identifiable with protected attributes (e.g., males and females) and therefore a decision process leveraging such information might lead to fairness issues\,\cite{mehrabi2021surveyFairness}. 
Therefore, the calculation of the background samples must take these issues into account to represent these groups fairly.
For datasets including a protected attribute where fairness issues may easily arise (i.e. HeartRisk and Student), we have applied K-means on the samples that identify males and females separately, to obtain two male samples and two female samples each as background samples.



\subsection{Iterative Dataset Weighting (IDW)}
\label{sec:contributo_IDW}

This first methodology combines all the explanations from an initial dataset $D_0$ into a new dataset $D_1$ that has the same shape of $D_0$, but whose values are altered by a matrix of importance values $|E_0|$.
Row $i$ of matrix $E_0$ is obtained as the SHAP scores for sample $i$ in $D_0$ as classified by $f_0$. 
Thus, each entry $E_0[i,j]$ represents the importance that $f_0$ gives to feature $j$ when classifying sample $i$.
The intuition of IDW is that a new black-box model $h_1$ trained on $D_1$ trains over a dataset where only the relevant features are preserved, while non-essential features (that will have close-to-zero explanation scores) will be largely suppressed. This approach aims at training a new model that retains most of the initial classification accuracy, but that relies only on a subset of features that are locally relevant.

\begin{figure}[ht]
    \centering
    
    \includegraphics[width=0.5\columnwidth]{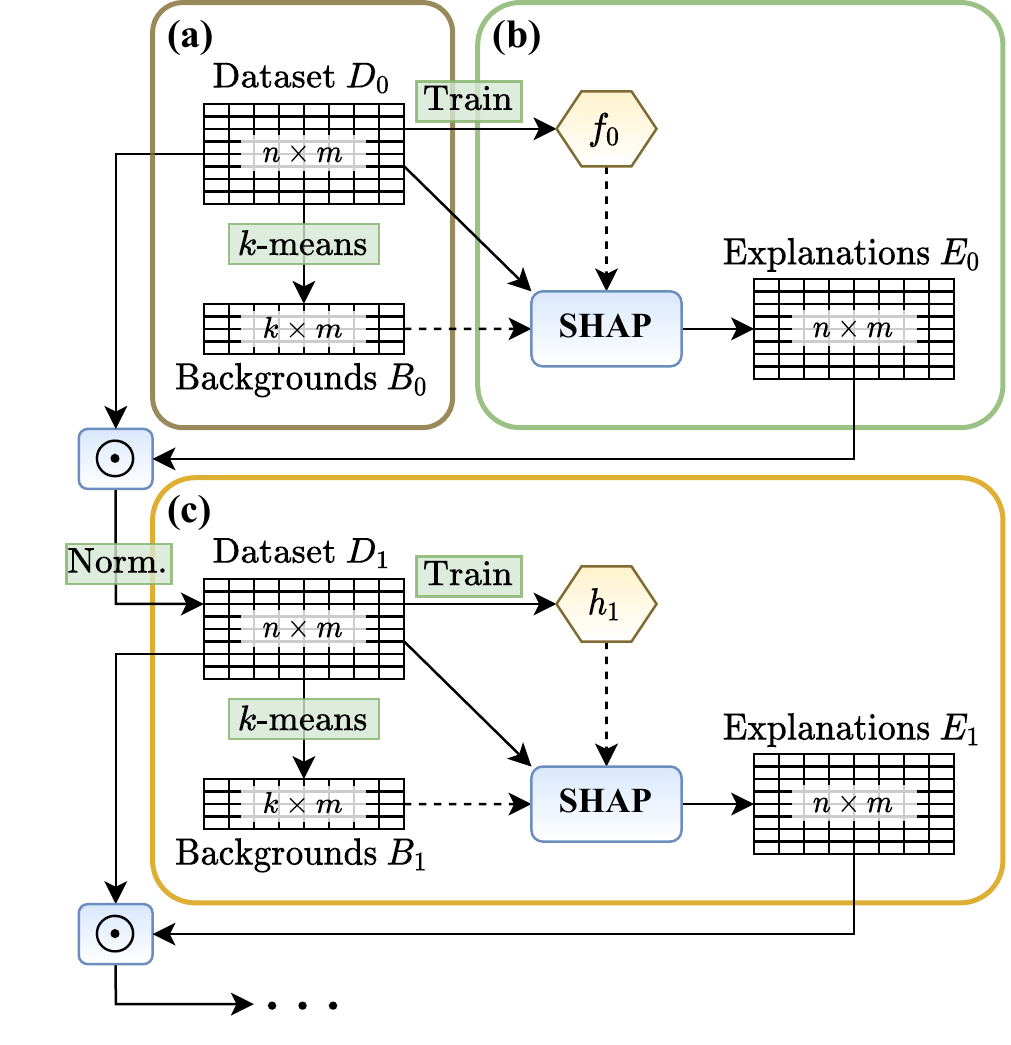}
    \caption{Workflow of IDW methodology.}
    \label{fig:IDW_SCHEMA}
\end{figure}

Fig~\ref{fig:IDW_SCHEMA} illustrates the iterative IDW workflow, following these steps:
\begin{enumerate}[label = \textbf{(\alph*)}]
    \item From dataset $D_0$ (preprocessed as described in \textit{Section \ref{sec:dataset}}) the background sample set $B_0$ is computed, and a black-box model $f_0$ is trained.
    
    \item SHAP is used to compute the explanations of every training sample in $D_0$, thus resulting in a matrix of SHAP scores $E_0$, where each entry $|E_0[i,j]|$ is the importance of feature $j$ for sample $i$ in $D_0$.
    
    
    \item A new dataset $D_1$ is obtained by combining the matrix $|E_0|$ with the initial dataset matrix $D_0$ using the element-wise Hadamard product, i.e.
    \begin{equation}\label{eq:D1fromD0}
        D_1 = |E_{0}| \odot D_{0} 
    \end{equation}
    and then rescaled using each feature's standard deviation, to keep columns as z-scores. 
    The process is performed for both the train and the test sets,
\end{enumerate}
Once $D_1$ is generated, the entire process can be repeated iteratively, generating a new background sample set $B_1$, a new black-box model $h_1$, a new SHAP scores matrix $E_1$, and a new combined dataset $D_2$.
The evaluation of IDW is therefore carried out on multiple $h_i$ models, obtained after $i$ iterations of the workflow in Fig~\ref{fig:IDW_SCHEMA}.
The goal is to check how the streamlined models $h_i$ compare against the initial model $f_0$.
We remark that every iteration requires to generate SHAP explanations, which is a computationally expensive operation.

\subsection{Targeted Replacement Values (TRV)}
\label{sec:contributo_TRV}

This second methodology starts with identifying a set of \textit{replacement values $R$}, one for each feature, and then these values are used to replace all non-relevant features in the original dataset $D_0$.
In order to decide which feature values are relevant, TRV sets a cognitive budget of at most $K$ relevant features to be retained in each sample of $D_0$.
Therefore, TRV differs from IDW by the strategy that it uses to mask the dataset matrix entries: IDW affects all feature values with a linear transformation, whereas TRV impacts only $m - K$ features (for each data point), mapping them to \textit{replaced} values.

\begin{figure}[ht]
    \centering
    \includegraphics[width=0.5\columnwidth]{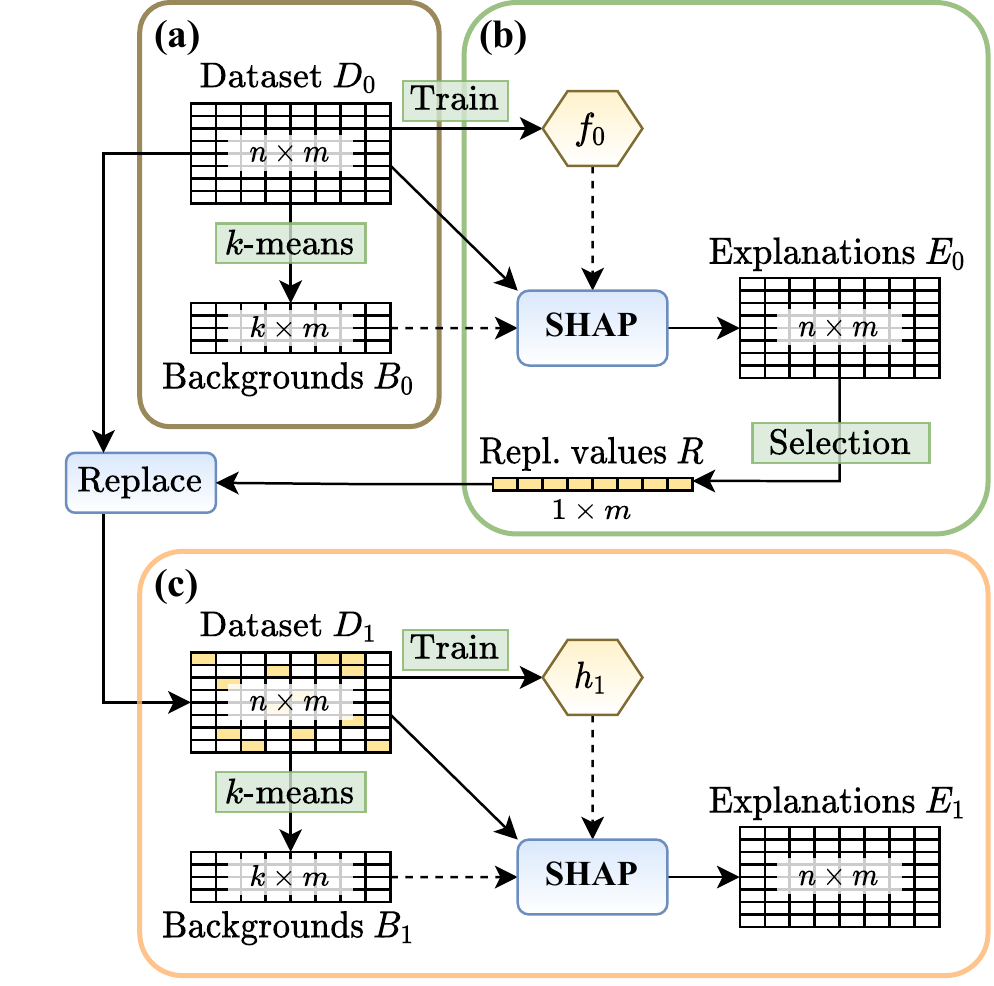}
    \caption{Workflow of TRV methodology.}
    \label{fig:TRV_SCHEMA}
\end{figure}

Fig~\ref{fig:TRV_SCHEMA} depicts the TRV workflow, which can be summarized in these steps:
\begin{enumerate}[label = \textbf{(\alph*)}]
    \item From dataset $D_0$ (preprocessed as described in \textit{Section \ref{sec:dataset}}) the background sample set $B_0$ is computed, and a black-box model $f_0$ is trained.
    
    \item SHAP is used to compute the matrix of SHAP scores $E_0$. 
        Using this matrix, the vector of replacement values $R$ is obtained as follows.
        For categorical features, the replacement value is always $0$, as we exploit the fact that these features were one-hot encoded using the $\{-1, 1\}$ values.
        For numerical features, we first set a threshold $\epsilon$, and then the replacement value for that feature is computed as the median value of all values in the dataset $D_0$ for that feature that have a corresponding absolute SHAP score in $E_0$ smaller than $\epsilon$,
        i.e.
        \begin{equation}
            R[j] = \mathrm{median}\bigl( D_0[C, j] \bigr)
        \end{equation}
        with $C$ being the set of rows indices $\bigl\{ i ~\big|~ |E_0|[i,j] < \epsilon \bigr\}$

    \item A new dataset $D_1$ is generated by keeping, for every sample, the top $K$ features \textit{by importance} (i.e. those that have the highest absolute SHAP score in $E_0$), and replacing all other features with their corresponding replacement value in $R$.
    This new masked dataset $D_1$ is then used to train a new black-box model $h_1$.
\end{enumerate}

TRV is not an iterative process, unlike IDW,  and the number of masked features is controlled by a parameter $K$.
Moreover, replacement values of numerical features do not have to be zeros, while in IDW the zero is implicitly the value that encodes the concept of non-relevance for every feature.

\FloatBarrier

\section{Experimental Results}
\label{sec:res}

In this section we test how the new streamlined models obtained by IDW and TRV compare with the original black-box model $f_0$.


\subsection{Model Architectures and Training}
\label{sec:bbox}

We have run experiments with a number of different classification algorithms - however, for the sake of simplicity, in this section we describe the results obtained with a single black-box model architecture, a Multi-Layer Perceptron (MLP).
Therefore the model architecture used for both $f_0$ and all $h_i$ is an MLP with one hidden layer. 
The hidden layer has a number of neurons equal to the number of features $m$ in the dataset on which the MLP is trained, and we set \textit{ReLU}s as the activation function for the input and hidden layers and \textit{sigmoid} as the activation function for the output layer. 
Furthermore, we used a learning rate of 0.001, \textit{binary cross-entropy} as loss function, and an early stopping criterion using a patience value of 10.


The results produced by IDW and TRV are further compared to standard regularized models (L1, L2 and L12) computed using the regularization factor set to 0.0001.
We do so to provide a baseline for induced model sparsity~\cite{henderson2021improving}.

\subsection{Datasets}
\label{sec:dataset}

We have tested our methodologies on several tabular datasets, and we report our results for: \emph{Cancer}~\cite{Dua:2019}, \emph{HeartRisk}~\cite{heart}, \emph{Kidney}~\cite{Dua:2019}, \emph{Student}~\cite{article}.

\begin{table}[!htb]
    \centering\setlength\tabcolsep{2pt}
    \scalebox{1.2}{
        \begin{tabular}{|c|c|c|c|c|c|c|c|}
            \hline
            \thead{\textbf{Dataset} \\ \textbf{Name}} & 
             \thead{\textbf{Numerical} \\ \textbf{Features}} & \thead{\textbf{Categorical} \\ \textbf{Features}} & \thead{\textbf{Binary} \\ \textbf{Features}} & \thead{\textbf{Preprocessed} \\ \textbf{Features}} & \thead{\textbf{Missing} \\ \textbf{Values}} & \thead{\textbf{Classification}}  & 
            \thead{\textbf{Protected} \\ \textbf{Attribute} }\\ \hline
            Cancer & 30 & 0 & 0 & 30 & \xmark & Binary & \xmark \\ \hline
            HeartRisk & 9 & 0 & 6 & 15 & \cmark & Binary & \cmark \\ \hline
            Kidney & 11 & 3 & 10 & 37 & \cmark & Binary & \xmark  \\ \hline
            Student & 13 & 4 & 13 & 43 & \xmark & Binary & \cmark  \\ \hline
        \end{tabular}
    }
    \caption{Characteristics of the datasets used in the computational experiments.}

    \label{tab:dataset_review}
\end{table}

A summary of the dataset characteristics is shown in \textit{Table \ref{tab:dataset_review}}. For each dataset, \emph{Numerical Features} and \emph{Categorical Features} indicate the number of numerical and categorical features, respectively;
\emph{Preprocessed Features} indicates the number of features after the preprocessing phase;
\emph{Missing values} states the presence of data points with missing values in one or more features;
\emph{Classification} denotes the classification task for which the dataset is designed; and
\emph{Protected Attribute} indicates whether the dataset contains some features that could potentially cause fairness issues.
For simplicity, we only consider datasets for binary classification, albeit our work can naturally be extended to other supervised ML tasks.

\subsection{Evaluation Scores}
\label{sec:risultati_metriche}
Firstly, we quantified potential declines or improvements in model performances (if any) for the proposed methodologies. We relied on standard model accuracy\footnote{
Note that datasets were rebalanced, so accuracy does not suffer from any imbalance bias.}
to analyze the ability of the models to correctly classify the samples of the dataset: $Acc = \frac{\TP + \TN}{\TP + \TN + \FP + \FN}$ with $\TP$, $\TN$, $\FP$ and $\FN$ being the standard confusion matrix entries for a binary classifier.

For the datasets where two sensitive groups $A$ and $B$ were present, we also consider a set of fairness metrics~\cite{10.1145/3375627.3375808}.
We include such metrics because we want to check that our feature engineering does not introduce in the classification pipeline an alteration of the fairness performances. 
We evaluated the Positive Predictive Parity as $\PPR_D = \frac{\TP_A}{\TP_A + \FP_A} - \frac{\TP_B}{\TP_B + \FP_B}$; the Negative Predictive Parity  as $\NPR_D = \frac{\TN_A}{\TN_A + \FN_A} - \frac{\TN_B}{\TN_B + \FN_B}$; the False Positive Parity as $\FPR_D = \frac{\FP_A}{\TN_A + \FP_A} - \frac{\FP_B}{\TN_B + \FP_B}$ and the Equality of Opportunity as $\EO_D = \frac{\TP_A}{\TP_A + \FN_A} - \frac{\TP_B}{\TP_B + \FN_B}$.
These scores are calculated considering males and females as fairness sensitive groups, and are computed only on the tested datasets that present this distinction: HeartRisk and Student. While smaller values reflect fairer models, it has been proven that it is not possible to achieve perfect equality across all these metrics simultaneously~\cite{pleiss2017fairness}. %

In order to capture the ability of the proposed methodologies to produce explanations 
that are simpler and easier to understand than the ones produced by
the original model $f_0$, we introduce two new metrics, namely the \textit{Explanation Compactness Percentage (XCP)} and the \textit{Glocal Similarity (Sim)}.
XCP measures the fraction of features not involved in the explanations (i.e. SHAP scores with an absolute value lower than a threshold, set to $\epsilon {=} 0.01$). 
XCP quantifies the ability of a model to rely on a sparse set of features, and thus produce sparse local linear explanations. 
We introduce this score as a way to verify that the streamlined model learned the classification rules from a smaller set of features than the ones used by $f_0$.
An high XCP value means that the model produced sparser explanations, which are naturally more interpretable for human users.


A different aspect that we measure is the consistency of the produced explanations that are proposed to the user. 
The principle is that a user will be confused by the system if explanations vary significantly 
among the classified samples\,\cite{cartwright_2003}.
In fact, in the extreme case where all the local explanations are the same, the model would be fully replaced by a global white box implementing the local rules without any loss in accuracy. In such case, the user would face the simplest scenario where features have the same impact on every sample, making much easier interpreting the model logic. However, substituting a complex model with a simpler global one is not always doable, and therefore with this metric we aim at quantifying how well local explanations of the new models tend to be collectively aligned towards a global one.
The \textit{Glocal Similarity} measure ($\Sim$) is introduced to capture how similar are the local explanations to a common global explanation in terms of their respective sets of explaining features. 
Since a global explanation is typically used to have an understanding on the expected features driving the average model predictions, having local explanations that are significantly divergent from the globally-relevant features would cause an additional cognitive burden to the end user to reconciliate between counter-intuitive signals. 
The vector of global importance of all features is calculated averaging the local importance across all the $n$ data points:
\begin{equation}\label{eq:globalExpl}
        g = \frac{1}{n} \sum_{i = 1}^n |E[i]|
\end{equation}
To measure such consistency, we define the Glocal Similarity as one minus the Hamming distance between the set of relevant features in $|E[i]|$ and $g$.
In this way, we assign an higher similarity if both $g$ and $E[i]$ share similar sets of relevant features for the explanation.
To compute the Hamming distance between $|E[i]|$ and $g$ such vectors are first processed into binary vectors $\overline{E}[i]$ and $\overline{g}$ using a threshold $\epsilon$ i.e. the values of $|E[i]|$ and $g$ greater than $\epsilon$ are represented by 1, the remaining by 0 respectively in $\overline{E}[i]$ and $\overline{g}$.
Therefore, the Glocal similarity for the sample $i$ is computed as 
\begin{equation}\label{eq:glocal}
        \Sim(i) = 1 - \frac{c(\overline{E}[i],\, \overline{g})}{m}
\end{equation}
where $c(a,b)$ counts the number of occurrences of elements with different values in $a$ and $b$, in the same positions.
In our experiments, we used a threshold value $\epsilon {=} 0.01$.
Note that an higher $\Sim$ value does not imply that the model producing the explanation is better than another, but just that it is more consistent in identifying the relevant features for different samples. For instance, a linear model would result in a perfect $\Sim$ score. 
Given our stated Research Question~\eqref{eq:RQ}, we focus on measuring the coherence among multiple explanations.

\subsection{Iterative Dataset Weighting Results}
\label{sec:risultati_IDW}

\begin{figure}
    \centering
    \includegraphics[width = .65\linewidth]{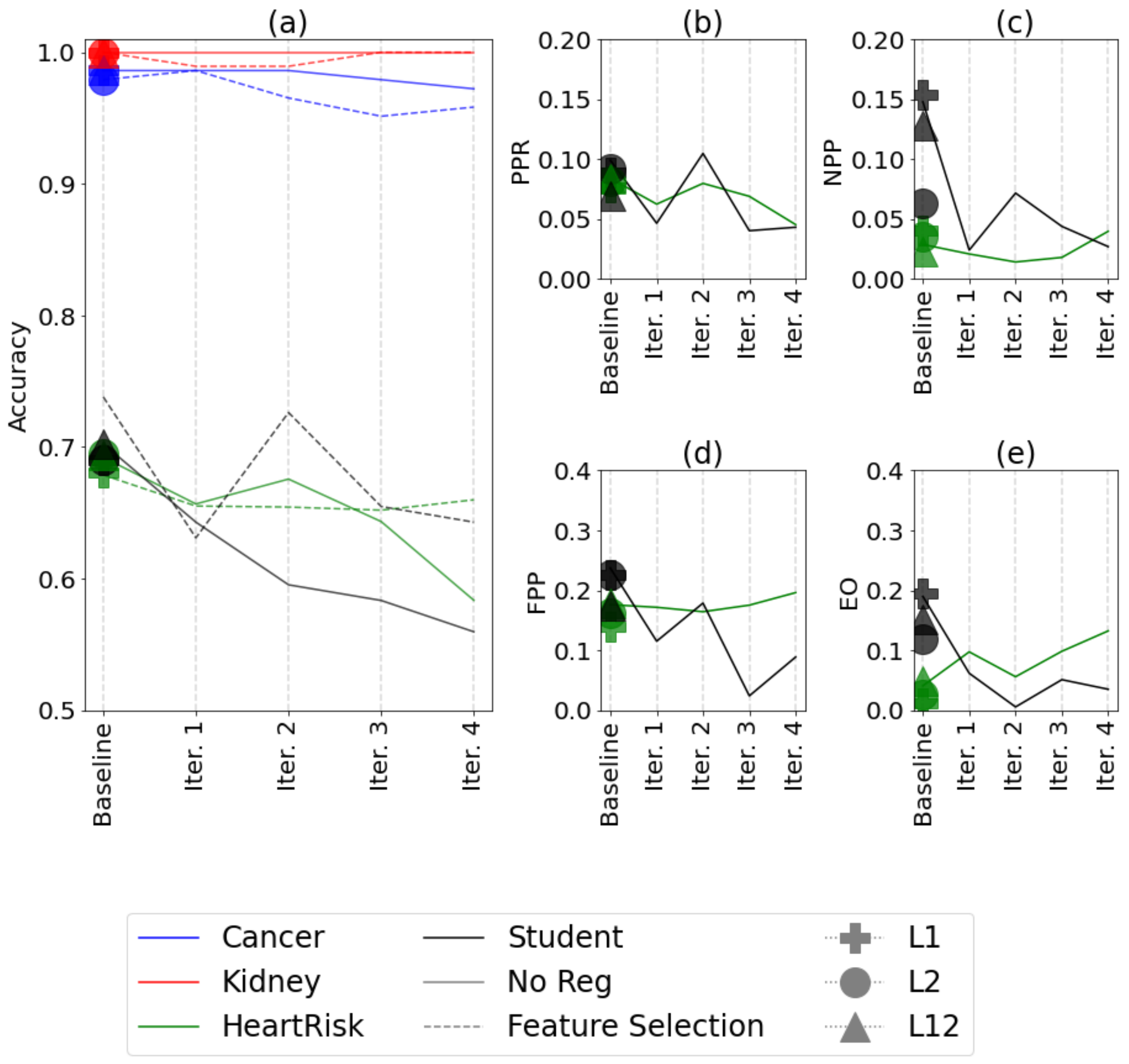}
    \caption{Accuracy (a) and absolute value of fairness metrics (b-e)
    for the baseline (without regularization)) and the IDW streamlined models, computed on the test sets. Each color identifies a dataset.}
    \label{fig:acc-fair:IDW}
\end{figure}

\begin{figure}
    \centering
    \includegraphics[width = .5\linewidth]{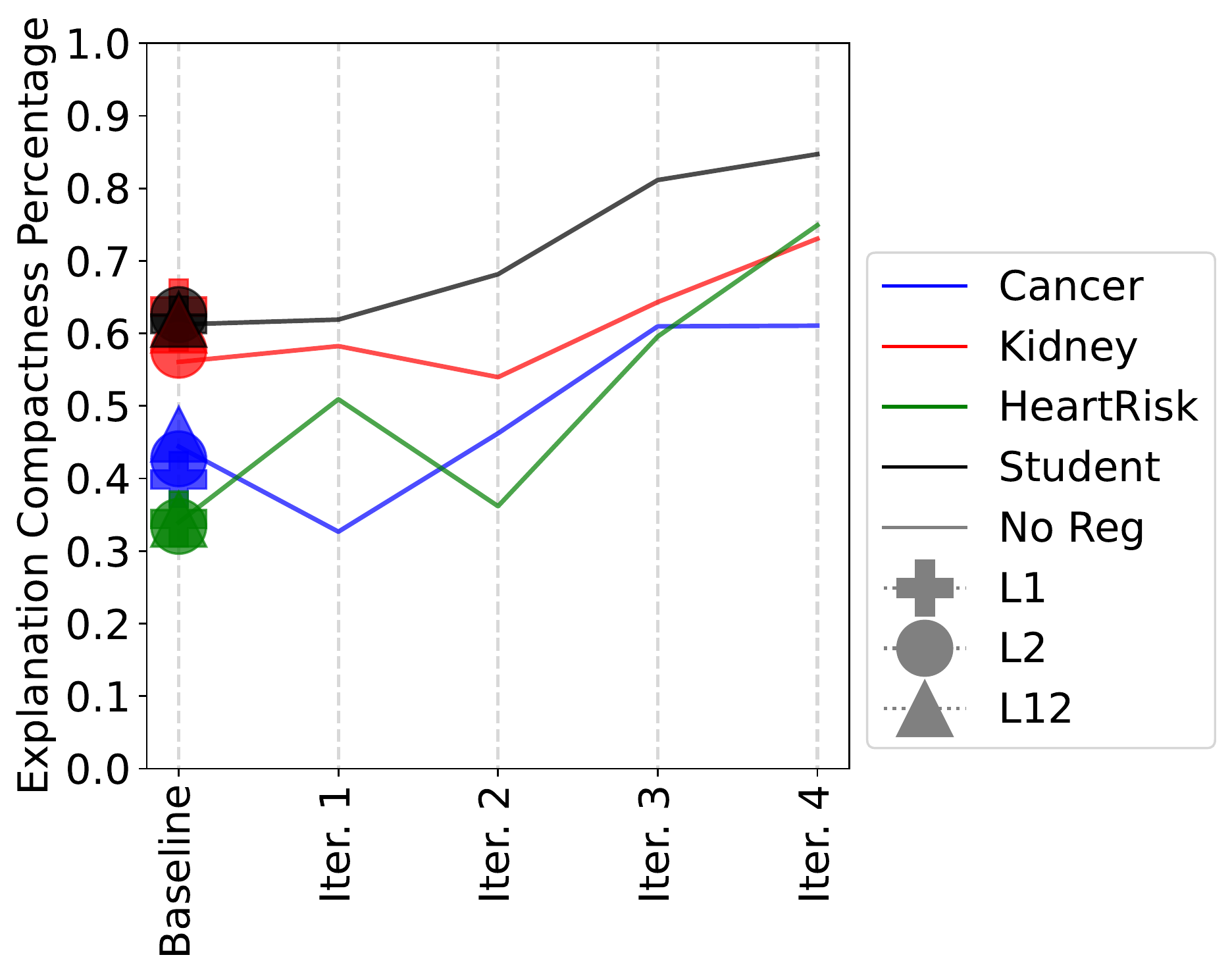}
    \caption{XCP of the baseline (without regularization) and of the streamlined models computed using IDW, on the test sets.}
    \label{fig:XCP:IDW}
\end{figure}

\begin{figure}
    \centering
    \includegraphics[width = .65\linewidth]{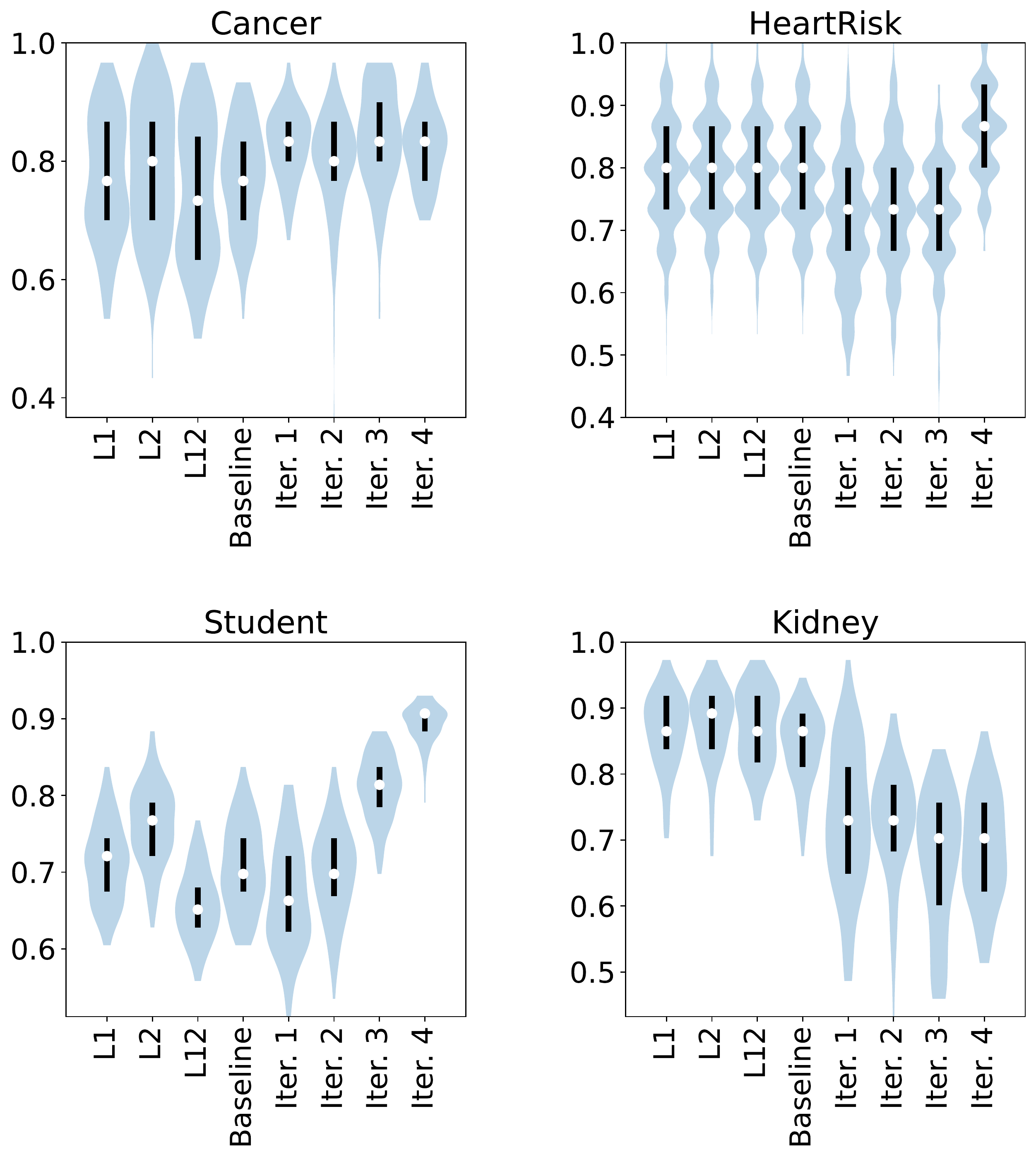}
    \caption{
    Distribution of the $\Sim$ scores (y-axis) of the streamlined IDW models and of the baseline (without regularization), computed on the explanations of the text set.}
    \label{fig:sim:IDW}
\end{figure}

\begin{figure}
    \centering
    \includegraphics[width = \linewidth]{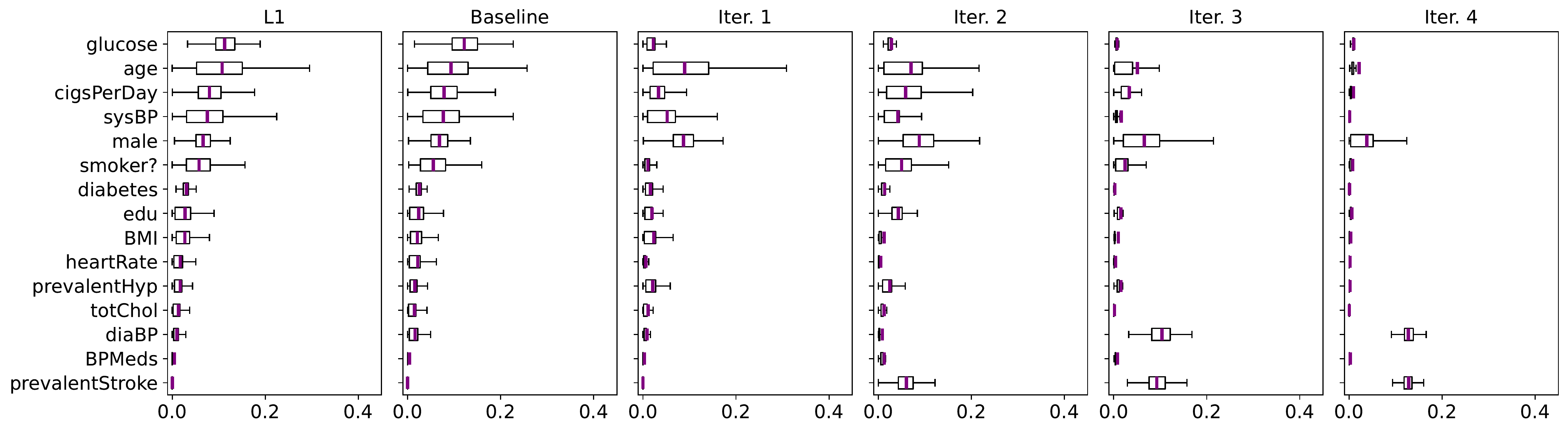}
    \caption{
    Box-plots of the absolute SHAP scores (x-axis) computed on the explanations obtained by the baseline and the IDW streamlined models on the test set, for the HeartRisk dataset.
    Each row represents the distribution of a feature in the HeartRisk dataset.}
    \label{fig:expl:IDW}
\end{figure}

In this section the results produced using IDW are illustrated up to 4 iterations.
Fig.~\ref{fig:acc-fair:IDW}(a) reports the accuracy and the absolute fairness metrics for the baseline model (with and without regularization), as well as the streamlined IDW models up to the fourth iteration.
In Fig.~\ref{fig:acc-fair:IDW}(a) we can observe that on the datasets Cancer and Kidney IDW caused only negligible decreases in accuracy, whereas HeartRisk and Student suffered a greater loss of accuracy, as the iterations progress. 
The dotted line reports the accuracy of another model where a global feature selection~\footnote{\href{https://scikit-learn.org/stable/modules/generated/sklearn.feature_selection.SelectKBest.html}{SelectKBest}, from Scikit-learn, with default parameters.} is applied, keeping only the top-$K$ relevant features, with $K$ being the average number of features with importance above $\epsilon$. Global feature selection seems to have comparable performances than IDW.
This shows that, in general, one should expect some accuracy loss when the model is streamlined, as the new model is focusing on a much smaller set of features to provide its decisions.
The absolute fairness metrics in Fig.~\ref{fig:acc-fair:IDW}(b-e) show that these metrics were not significantly impacted by IDW, even if some variations are observed.

Second, we measure the XCP of the streamlined IDW models $h_1 \ldots h_4$ compared to that of the baseline model $f_0$ and, additionally, to that of the regularized baseline models.
Fig.~\ref{fig:XCP:IDW} shows the XCP scores for the baseline and for IDW at each iteration. 
In all the tested datasets, the compactness of explanations shows an improvement trend using IDW. This trend shows an increase in the number of features that are considered not relevant in the explanations, for higher iterations.
Moreover, the regularized models have only marginal improvements in explanation sparsity compared to the streamlined IDW models.

Fig.~\ref{fig:sim:IDW} shows the distributions of the \textit{glocal similarity} scores computed at each iteration of IDW.
The results show that IDW does not always produce explanations that are more similar to a common global explanation. Only 2 datasets (Cancer and Student) appear to have a general improvement trend of the $\Sim$ score, while HeartRisk gets an improvement only at the last iteration.
Hence one desirable property of model streamlining is not achieved consistently.

To better understand the streamlining effect of IDW, Fig.~\ref{fig:expl:IDW} shows the distributions of the absolute SHAP scores for the explanation matrix $E_0$ and each successive matrix $E_1 \ldots E_4$. For the sake of completeness, we also report the result for the explanations generated by a L1 regularized model.
The distribution of the absolute SHAP scores for the baseline case has values scattered for almost all features. 
As the IDW iterations proceed, the model is forced to concentrate on an increasingly smaller set of features. 
Model $h_4$ gives importance to about 4 features only.
This shows that the feature engineering achieves the goal of explanation streamlining. Therefore we have achieved a model that locally considers only very few features, at the price of a decreased classification accuracy.
We clarify that this small set of features is defined per-sample, and not just a projection of the whole dataset. Thus the streamlined model differs from a model that is constructed by an aggressive simplification of the problem through a global feature selection phase.

\subsection{Targeted Replacement Values Results}
\label{sec:risultati_TRV}

\begin{figure}
\centering
  \centering
    \includegraphics[width =.65\linewidth]{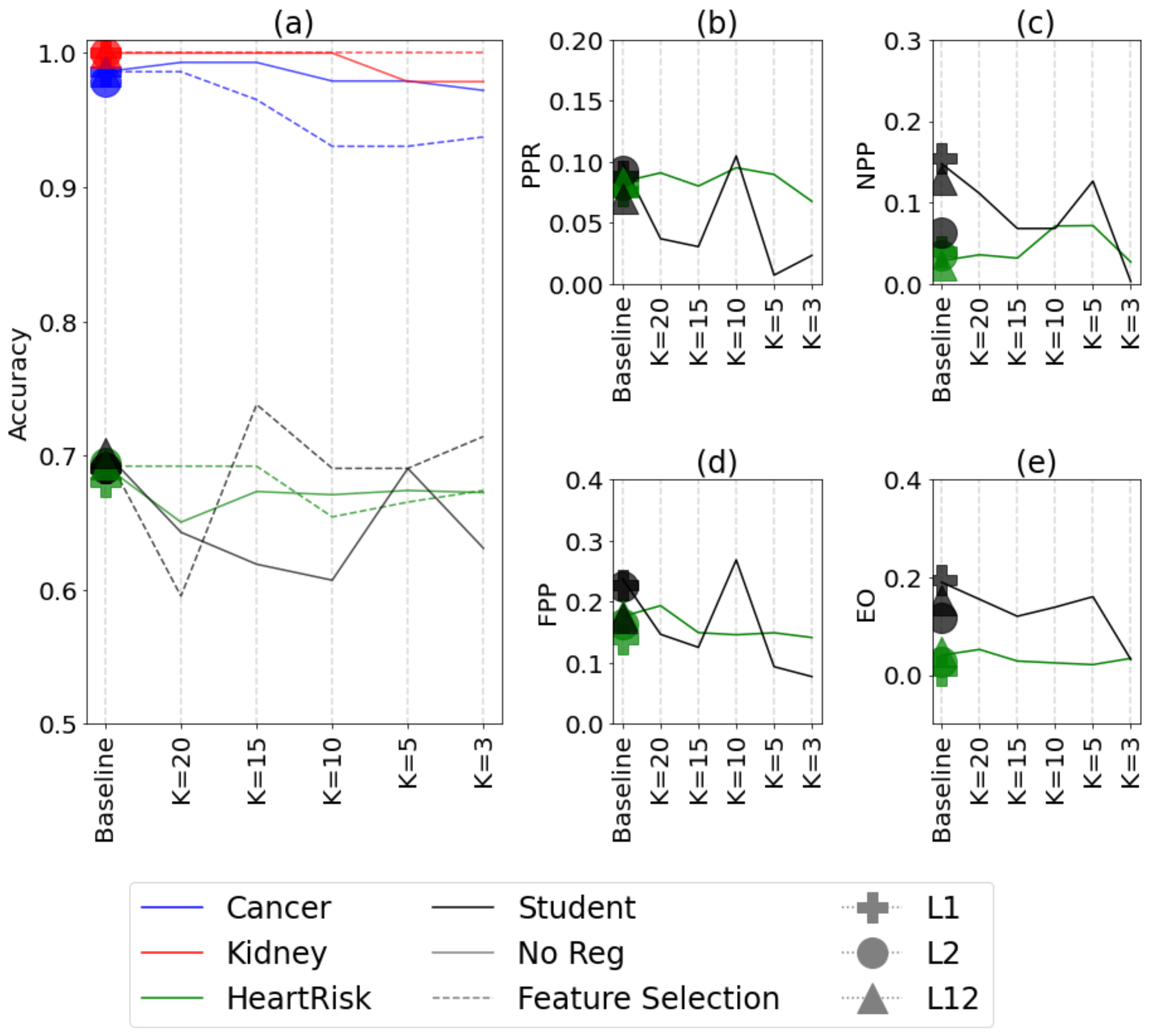}
    \caption{Accuracy (a) and absolute value of fairness metrics (b-e) for the baseline (without regularization)) and the TRV streamlined models for different $K$ values, computed on the test sets. Each color identifies a dataset.}
    \label{fig:acc-fair:TRV}
\end{figure}

\begin{figure}
    \centering
    \includegraphics[width =.5\linewidth]{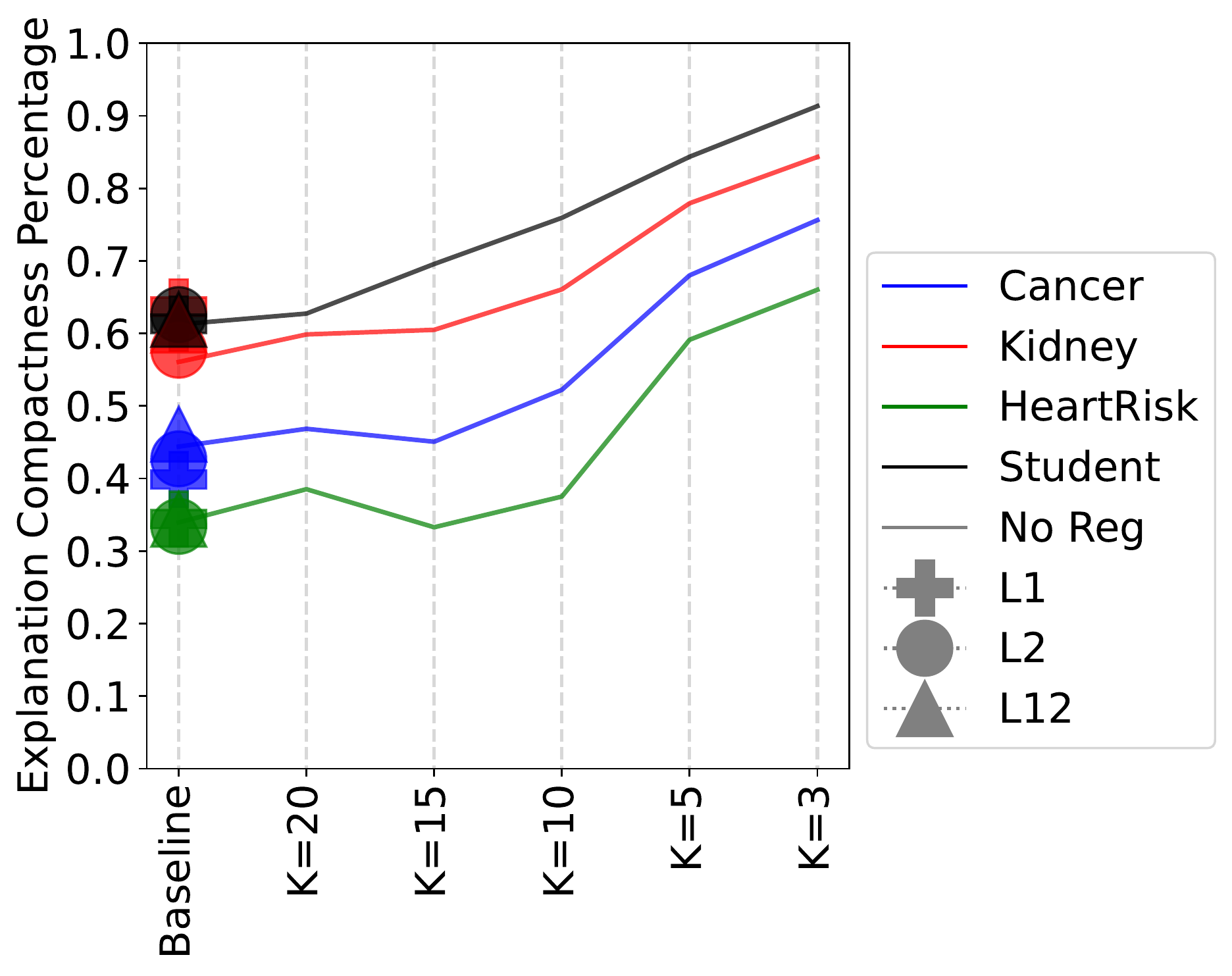}
    \caption{XCP of the baseline (without regularization) and of the streamlined models computed using TRV, on the test sets.}
    \label{fig:XCP:TRV}
\end{figure}

\begin{figure}
    \centering
    \includegraphics[width = .65\linewidth]{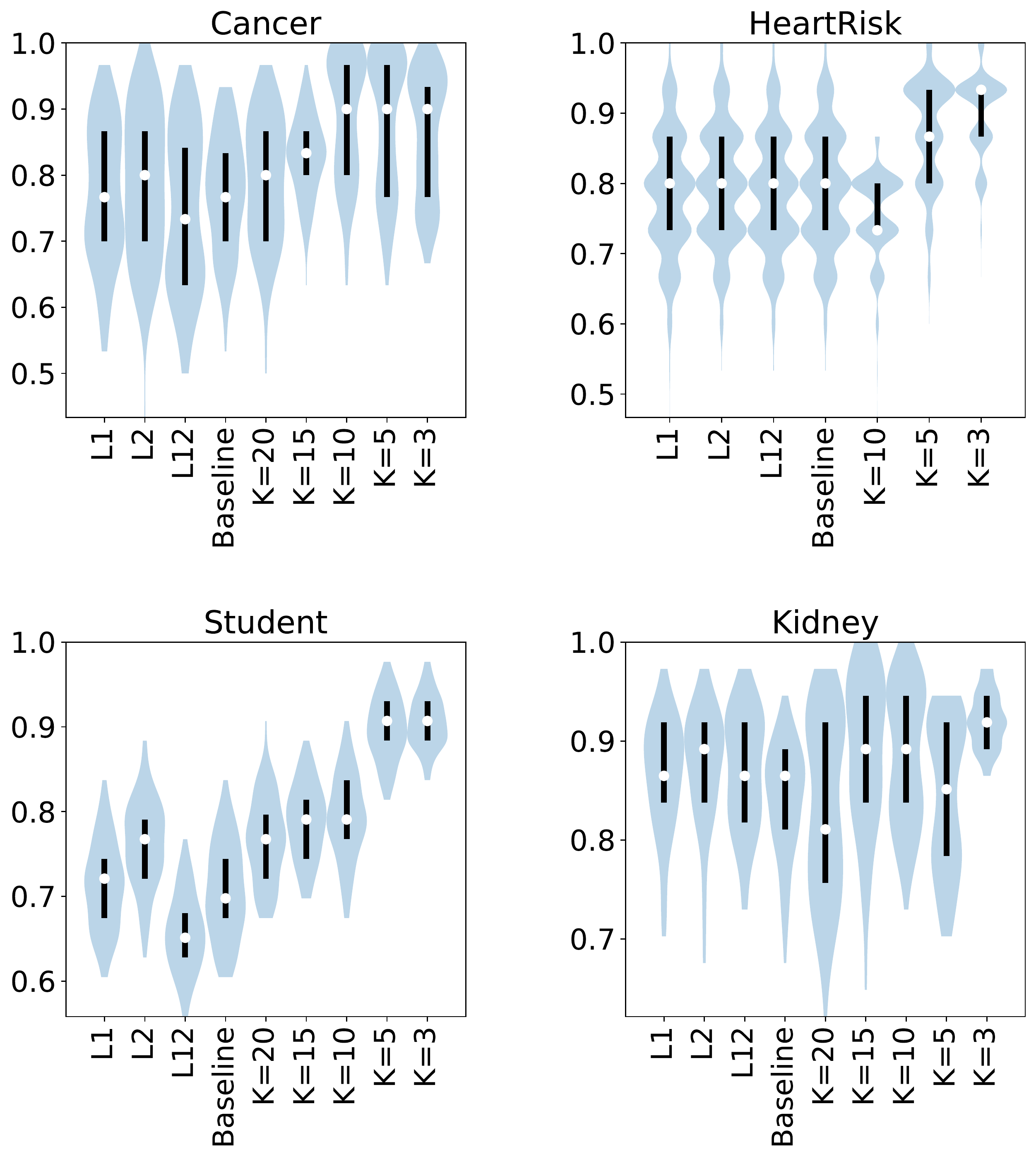}
    \caption{Distribution of the $\Sim$ scores (y-axis) of the streamlined TRV models and of the baseline (without regularization), computed on the explanations of the text set.}
    \label{fig:sim:TRV}
\end{figure}

\begin{figure}
    \centering
    \includegraphics[width = \linewidth]{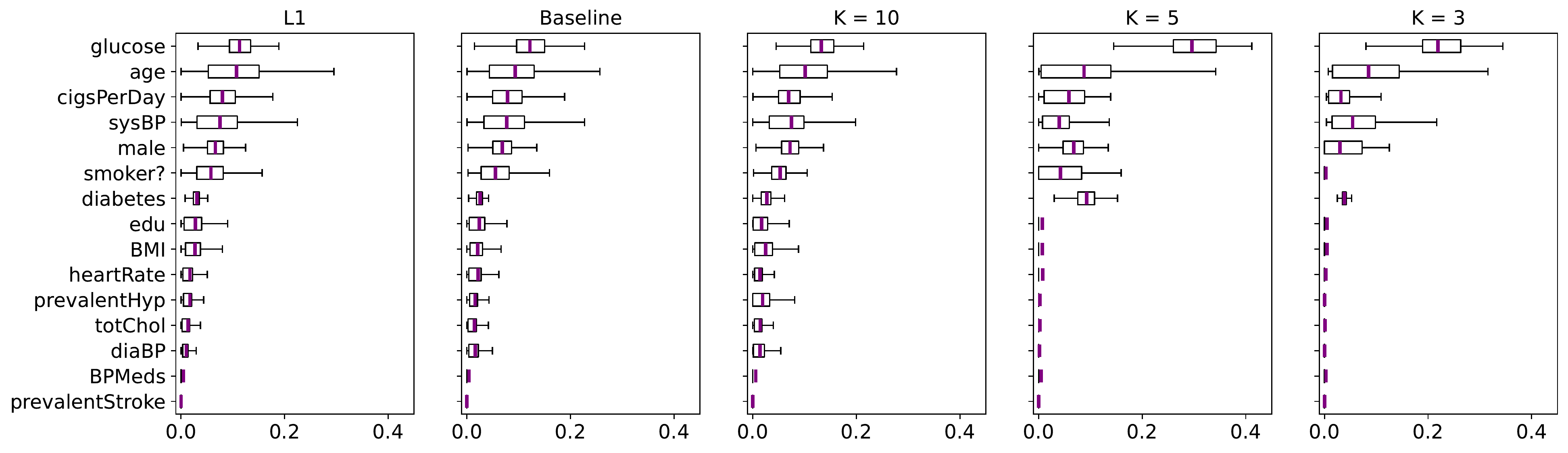}
    \caption{Box-plots of the absolute SHAP scores (x-axis) computed on the explanations obtained by the baseline and the TRV streamlined models on the test set, for the HeartRisk dataset.
    Each row represents the distribution of a feature in the HeartRisk dataset.}
    \label{fig:expl:TRV}
\end{figure}

In the TRV pipeline, all but the $K$ most important features (according to their explanation scores) are replaced with the \textit{replacement values} $R$ (See \textit{Section \ref{sec:contributo_TRV}}). 
Thus, the user interested in an explanation selects in advance the number $K$ of relevant features to get for every sample.
The values for $K$ used in our tests are $3$, $5$, $10$, $15$ and $20$.
Lower values for $K$ are particularly relevant, following the observations in~\cite{kouki2019personalized} where the authors report that the average number of features before a user loses its interest is $4.32$.

TRV determines which importance values denote a feature being relevant or not by comparison with a threshold $\epsilon$. The results were obtained using $\epsilon{=}0.01$ for all the tested datasets.
Such threshold requires some tuning. In our experiments we observed that for these four datasets the selected value is a good indicator for low SHAP scores. However, for larger datasets with hundreds of features, a smaller threshold could be required.

For TRV, we followed the same experimental setup already outlined for IDW.
Fig.~\ref{fig:acc-fair:TRV} shows the accuracy scores obtained by the baseline model (without regularizations) compared with the streamlined models generated using TRV for different $K$ values.
In general TRV seems to preserve better the classification accuracy that IDW, with only one dataset among the tested ones (Student) showing some relevant degradation.
Moreover, fairness metrics do not appear to have been impacted significantly by the feature engineering process of TRV.
As before, the dotted represents the accuracy of a model obtained after a global feature selection keeping only $K$ features. Also in this case, the global feature selection appears to have has comparable results.

Fig.~\ref{fig:XCP:TRV} shows the XCP scores for increasing values of $K$ across all the tested datasets. 
In this case, TRV shows a significant capacity of increasing the compactness of the explanations obtained from the streamlined models. 
Recall that the feature engineering of TRV alters $D_1$, while the XCP scores are computed on the explanation matrix $E_1$, i.e. it is not automatic that replacing $m-K$ values results in explanations with $K$ relevant SHAP values. 
Since gains in the XCP scores do not correspond to equally large decreases in accuracy, the experiments suggest a more consistent efficacy of this methodology over IDW in preserving the accuracy while making the model more consistent in predicting the class using a smaller set of features.

Fig.~\ref{fig:sim:TRV} shows the $\Sim$ scores for the explanations of the baseline model, the regularized models, and the models obtained using TRV.
Also this score hints that explanations tend to share the same set of features for smaller values of $K$, thus resulting in explanations that are both compact and more consistent to a single global explanation.
This effect seems to also be confirmed when observing the SHAP score distributions in Fig.~\ref{fig:expl:TRV} (for the HeartRisk dataset).
In that case, the global explanation involves a slightly larger set of features than IDW (6 for the $K=3$ case),  even if models will see $m-K$ replaced values for every input sample. But that set allows the model to keep a better accuracy, even if explanations will focus on slightly different feature sets among different samples.

\FloatBarrier

\section{Conclusions}
\label{sec:conclusions}
By definition, algorithm-generated explanation of ML models have to be understandable by humans. This concept, tightly coupled with the amount of information included in an explanation, has been referred to as {\em comprehensibility}~\cite{bib:guidotti}, {\em conciseness}~\cite{bib:LEAF}, and {\em compactness}~\cite{eval_xai} - amongst others.
Local linear explanations provide an insight on how a black-box classifier weights its input features to achieve the classification outcome of a given sample. 
Knowing this information, a streamlined model can be trained to focus more on the locally-relevant features, thus inducing this new model to concentrate on a (locally) small input set. 
Such system allows to design models that may still achieve reasonable accuracy, while their classification depends only on a smaller set of variables, at the cost of an expensive data engineering step.
The small set is not fixed (as in a global feature selection) but depends on the explained sample.
In this paper we reported the results on two experimental setups, where the streamlined models are obtained from two different data engineering processes, IDW and TRV.
These two setups learn from a combination of the initial data with the Shapley scores computed by a baseline model.

Experimental results show that the streamlined models were able to keep a competitive accuracy (especially for TRV), while at the same time being able to focus on a smaller set of input features. 
The proposed methods apply a logic that may look similar to a feature selection, but where the set of features is not global. A comparison with a standard global feature selection (reported in Fig.\ref{fig:acc-fair:IDW} and \ref{fig:acc-fair:TRV}) showed that, despite fluctuations in the results, the proposed techniques looks generally comparable; a thorough analysis is needed to make a full assessment.
Fairness metrics were also considered, in order to ensure that the resulting pipeline did not significantly harm the model fairness.
Moreover, the results outperform global feature selection approaches, since a form of locality is retained (i.e. every explanation is local and compact at the same time, but there is no single set of globally selected features), as well as explanations from regularized models, that were consistently less compact.

These experiments shows that it is possible to benefit from an explanation-in-the-loop approach. 
A baseline model provides hints to which features are more interesting at the single data point level, and this information is combined as input for a streamlined model. 
Thus, the final decision can be taken on the basis of a compact input set induced by the SHAP explanation scores.
Such compactness is a desirable property since this information better corresponds to the human intuition behind a sample explanation.

In the future, we plan to develop a user research on a real setting where decision makers assisted by ML models take advantage of post-hoc explanations during the process. In such setting we aim at measuring how much the streamlined models are able to increase trust, adoption and transparency in the whole decision making process.

A follow up study will be devoted the streamlining concept on different learning tasks such as regressions or image classification. Furthermore, following the very same intuition, we are interested in devising novel methodologies to include different types of explanations in the learning process, such as counterfactual explanations or motif-based explanations for graph classification tasks.

We believe that a framework that combines both explanations and learning to provide \textit{compact} and \textit{consistent} local explanations that are coherent with the classification process of a model is desirable, since it would overcome the limit of generating explanations a-posteriori without profiting from their information. 

\vspace{5pt}\noindent\textit{Reproducibility.}
All code and data used to perform the experiments in this paper are available in a GitHub repository\footnote{\url{https://github.com/FrancescoLomuscio/Explanations-in-the-loop/}}.

\FloatBarrier

\bibliographystyle{plain}
\bibliography{explanations_loop}

\end{document}